\documentclass{article}
\usepackage{arxiv}
\usepackage{times}
\usepackage{epsfig}
\usepackage{graphicx}
\usepackage{amsmath}
\usepackage{amssymb}
\usepackage{float}
\usepackage{bm}
\usepackage{algorithm}
\usepackage{pifont}
\newcommand{\cmark}{\ding{51}}%
\newcommand{\xmark}{\ding{55}}%
\usepackage[noend]{algpseudocode}
\usepackage{graphics}
\usepackage{caption}
\usepackage{subcaption}
\usepackage{xcolor}

\usepackage{hyperref}
\usepackage{microtype}

\usepackage{amssymb}
\usepackage{amsmath}
\usepackage{amsthm}
\usepackage{mathdots}
\usepackage{mathtools}

\usepackage{tensor}

\usepackage{urwchancal}
\DeclareFontFamily{OT1}{pzc}{}
\DeclareFontShape{OT1}{pzc}{m}{it}{<-> s * [1.10] pzcmi7t}{}
\DeclareMathAlphabet{\mathpzc}{OT1}{pzc}{m}{it}

\usepackage{graphicx}
\usepackage{ifpdf}
\ifpdf
\usepackage{epstopdf}
\PrependGraphicsExtensions{.svg}
\fi

\usepackage{xypic}

\usepackage{accents}
\makeatletter
\providecommand{\scirc}{%
    \hbox{\fontfamily{\rmdefault}\fontsize{0.4\dimexpr(\f@size pt)}{0}\selectfont{\raisebox{-0.52ex}[0ex][-0.52ex]{$\circ$}}}}

\makeatother

\mathchardef\mhyphen="2D

\newcommand{\publicationdetails}
{Under review}
\newcommand{\publicationversion}
{Preprint}

\begin{document}

\title{Iterative Optimisation with an Innovation CNN for Pose Refinement}

\author{
\href{https://orcid.org/0000-0001-2345-6789}{\includegraphics[scale=0.06]{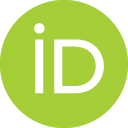}\hspace{1mm}
Gerard Kennedy}
\\
	Systems Theory and Robotics Group \\
	Australian National University \\
    ACT, 2601, Australia \\
	\texttt{Gerard.Kennedy@anu.edu.au} \\
	\And	\href{https://orcid.org/0000-0001-2345-6789}{\includegraphics[scale=0.06]{orcid.png}\hspace{1mm}
Zheyu Zhuang}
\\
	Systems Theory and Robotics Group \\
	Australian National University \\
    ACT, 2601, Australia \\
	\texttt{Zheyu.Zhuang@anu.edu.au} \\
	\And	\href{https://orcid.org/0000-0001-2345-6789}{\includegraphics[scale=0.06]{orcid.png}\hspace{1mm}
Xin Yu}
\\
	University of Technology \\
	NSW, 2007, Australia \\
	\texttt{Xin.Yu@uts.edu.au} \\
\And	\href{https://orcid.org/0000-0001-2345-6789}{\includegraphics[scale=0.06]{orcid.png}\hspace{1mm}
Robert Mahony}
\\
	Systems Theory and Robotics Group \\
	Australian National University \\
	ACT, 2601, Australia \\
	\texttt{Robert.Mahony@anu.edu.au} \\
}

\maketitle

\begin{abstract}
	Object pose estimation from a single RGB image is a challenging problem due to variable lighting conditions and viewpoint changes.
	The most accurate pose estimation networks implement pose refinement via reprojection of a known, textured 3D model, however, such methods cannot be applied without high quality 3D models of the observed objects.
	In this work we propose an approach, namely an Innovation CNN, to object pose estimation refinement that overcomes the requirement for reprojecting a textured 3D model.
	Our approach improves initial pose estimation progressively by applying the Innovation CNN iteratively in a stochastic gradient descent (SGD) framework.
	We evaluate our method on the popular LINEMOD and Occlusion LINEMOD datasets and obtain state-of-the-art performance on both datasets.
\end{abstract}


\section{Introduction \label{introduction}}
The task of object pose estimation involves estimating the 6D pose (translation and orientation) of a specified object relative to a specified reference frame.
This is an important problem in computer vision, and has a range of applications including robotic manipulation/grasping, and virtual/augmented reality.
Recent state-of-the-art RGB pose estimation algorithms use a two phase approach, initially providing a rough estimate of pose, and then using a second network to refine the pose estimate in order to obtain the desired performance \cite{ssd6d,dpod}.
The principle of refining an existing estimate of pose can be applied multiple times \cite{deepim,dpod,manhardt}.
Such an approach can be seen as analogous to iterative state refinement algorithms such as stochastic gradient descent (SGD) that are well established in the optimisation field \cite{fletcher}.

\begin{figure}[H]
    \hspace{-0.2in}
    \begin{subfigure}[b]{0.2\textwidth}
        \centering
        \includegraphics[width=1.1\textwidth]{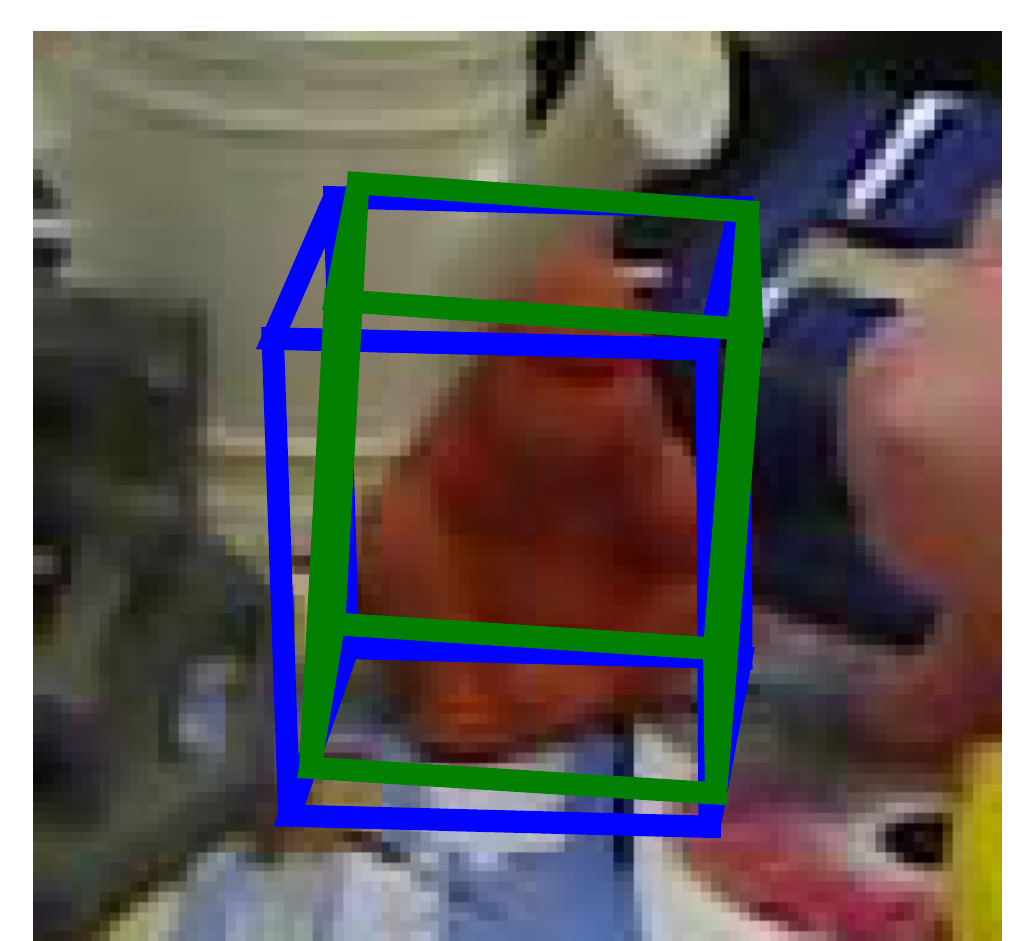}
        \caption{Initial Estimate}
    \end{subfigure}
    \hfill
    \begin{subfigure}[b]{0.2\textwidth}
        \includegraphics[width=1.1\textwidth]{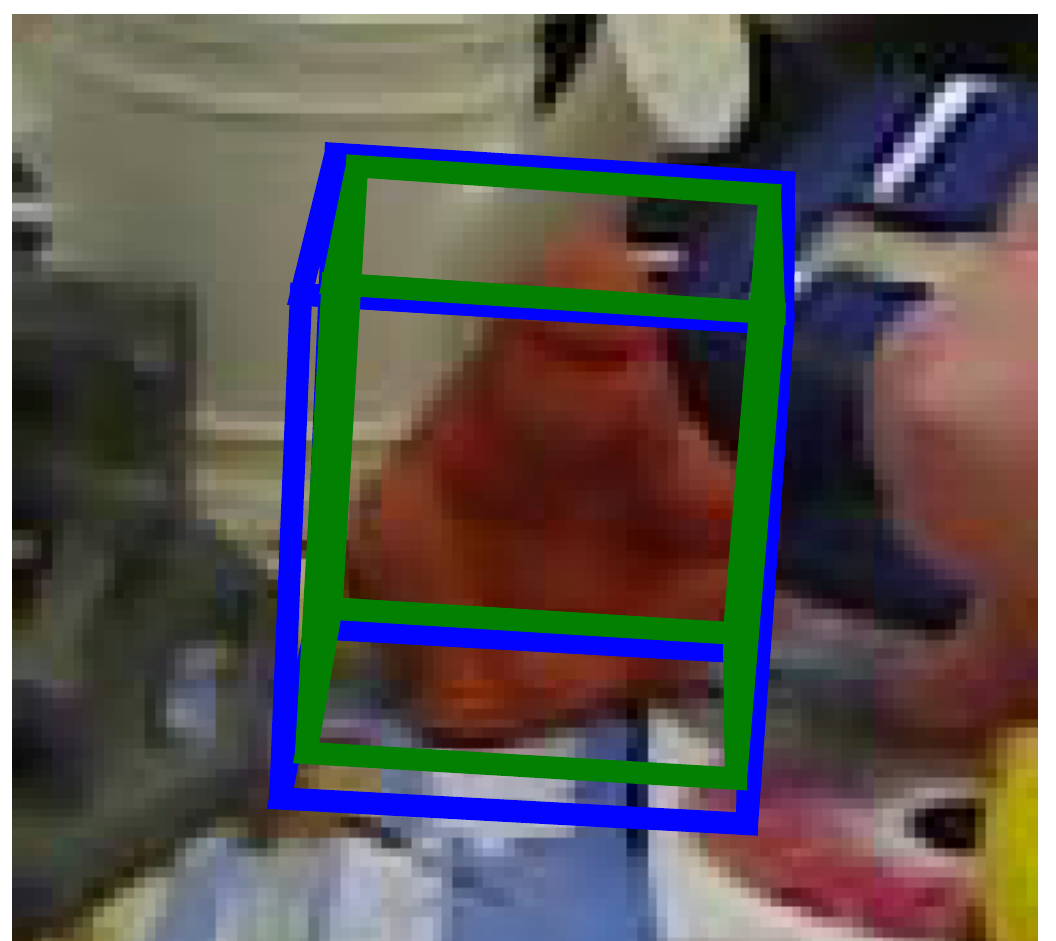}
        \caption{Iteration 1}
    \end{subfigure}
    \hfill
    \begin{subfigure}[b]{0.2\textwidth}
        \centering
        \includegraphics[width=1.1\textwidth]{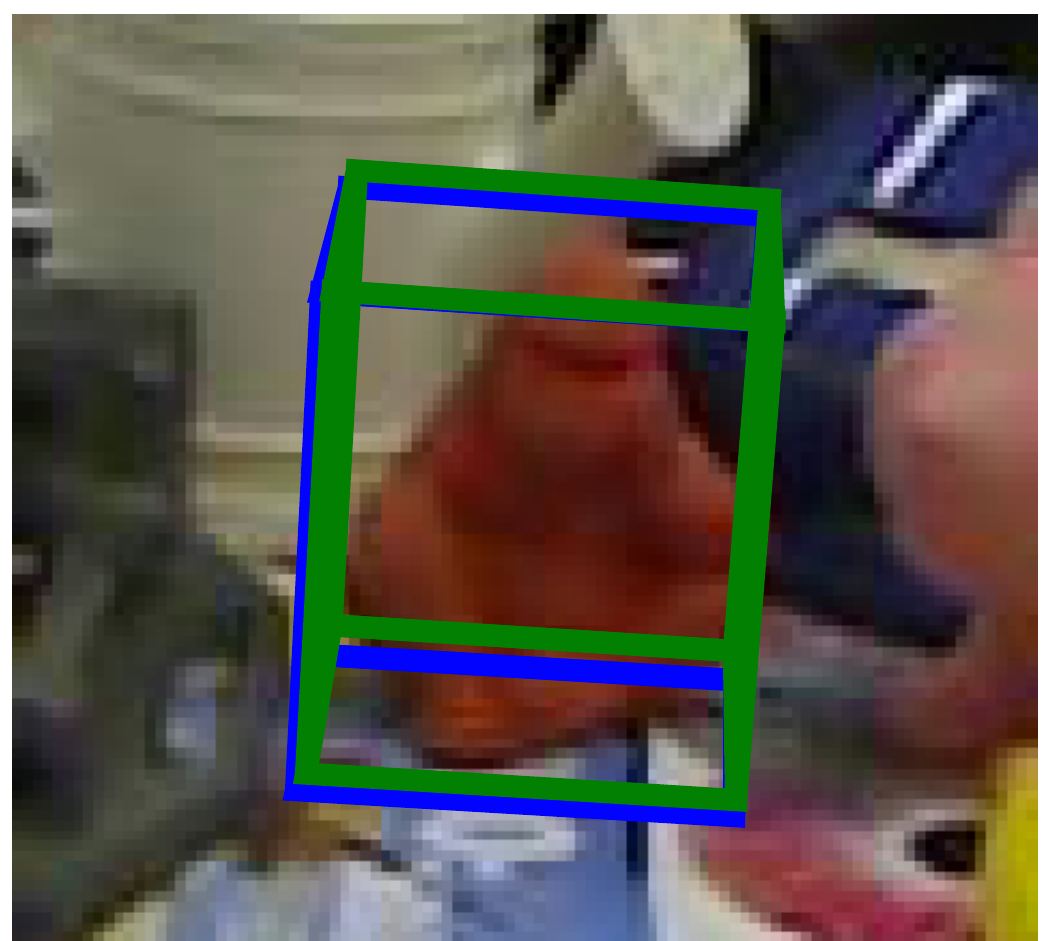}
        \caption{Iteration 2}
    \end{subfigure}
    \hfill
    \begin{subfigure}[b]{0.2\textwidth}
        \includegraphics[width=1.1\textwidth]{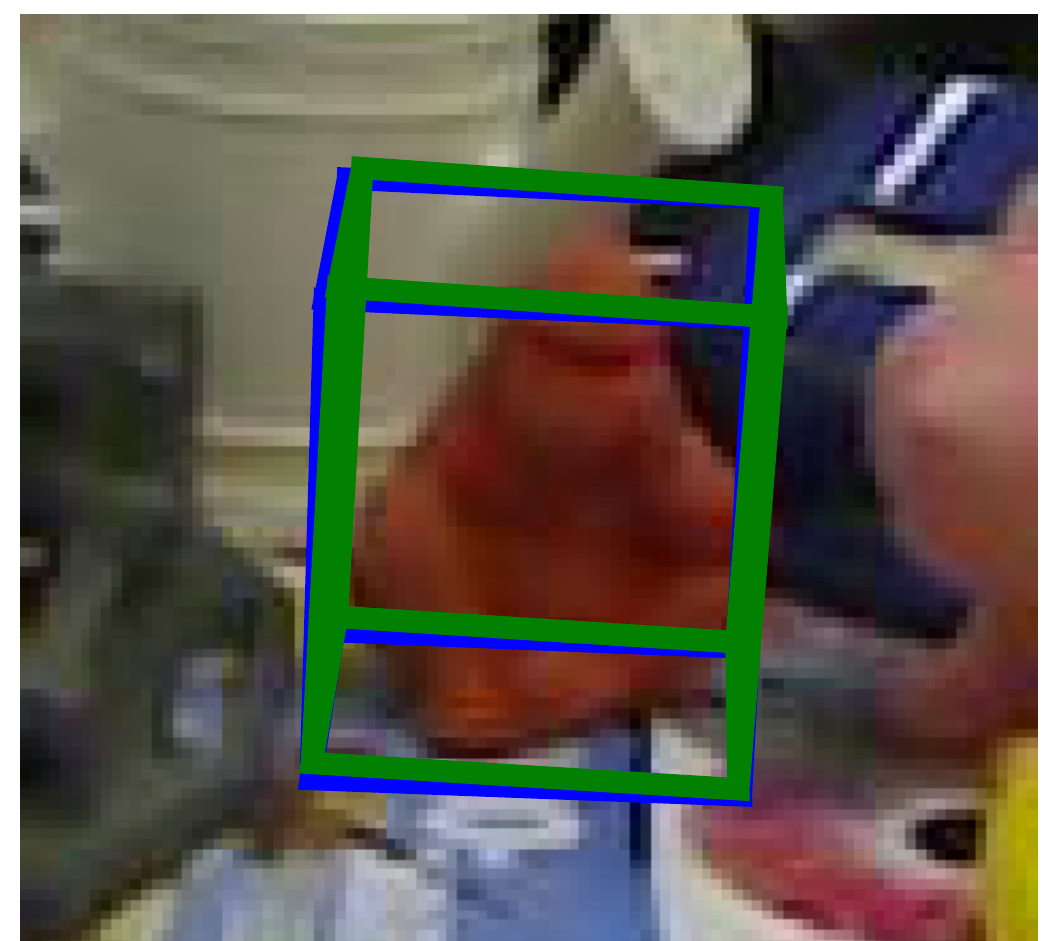}
        \caption{Final Estimate}
    \end{subfigure}
    \caption{Our proposed method applied to the Ape object of the \textbf{LINEMOD} dataset.
    {\color{green}Green} bounding boxes represent ground truth poses and {\color{blue}blue} boxes represent our results.}
	\label{fig:application}
\end{figure}

In this paper, we draw from the established principles of SGD and formulate a novel CNN architecture to implement an iterative refinement based on formal principles.
In doing so, we ease ease the requirement for the neural network to accurately estimate the pose in a single forward pass.
Specifically, we modify the loss function for a classical pose estimation network to estimate an update or refinement to an existing state, rather than estimating the state directly.
An autoencoder network is implemented in parallel with the modified pose estimation network architecture to encode the state estimate and inject this information into the modified pose estimation network using skip connections.
This overcomes the requirement to reconstruct an image from the state information in order for the state information to be injected.
The output of the modified pose estimation network is used as an SGD update for the state estimate.
Motivated by filtering theory we label this output the \textit{innovation}.
In the filtering field the innovation refers to the difference between the predicted and measured information.
We therefore call our algorithm the \textit{Innovation CNN}.
By formally recognising the role of the state estimate and the SGD update term, we are able to employ all the existing algorithms and insights on step-size choice and convergence analysis that have been developed for SGD.

We evaluate our approach on two widely used benchmarks for 6D object pose estimation, LINEMOD \cite{linemod} and Occlusion LINEMOD \cite{occlusionlinemod}.
Our approach refines initial pose estimates, leading to an average improvement of 3.69\% on LINEMOD and 3.31\% on Occlusion LINEMOD in terms of the widely-used ADD(-S) metric.
This leads us to achieve state-of-the-art results on both datasets.
We further demonstrate that our method is particularly effective at refining relatively poor initial estimates.
Specifically, we show relatively large improvement on the more difficult Occlusion LINEMOD dataset, along with particularly challenging objects of the LINEMOD dataset, such as ape (21.14\% improvement) and duck (10.73\% improvement) where the objects have poor surface texture.

In summary, the key contributions of this paper are:
\begin{itemize}
    \itemsep0em
    \item A novel framework that uses a CNN to estimate a state update term and utilise this to refine an initial estimate via SGD.
    \item A demonstration that this framework improves the results of an off-the-shelf object pose estimation network.
    \item State-of-the-art performance on the \textbf{LINEMOD} and \textbf{Occlusion LINEMOD} datasets\footnote{Recent work by Yu \emph{et al.} \cite{Yu2020} now outperforms our results.}.
\end{itemize}

\section{Related Work \label{related_work}}
Object pose estimation literature can be sub-divided into four general areas: end-to-end regression from image to pose, pose classification via a discretised pose space, regression to an intermediate representation such as keypoints, and pose refinement.
In this section we briefly review literature related to object pose estimation from a single RGB image based on these four categories.

\subsection{Pose Estimation}
\textbf{End-to-End Regression:}
End-to-end pose regression from a single image is a highly challenging task due to the nonlinearity of the space of rigid 3D rotations $SO(3)$.
This is addressed in \cite{posenet} by separating and carefully balancing translation and rotation terms in the loss function, although this is for camera pose estimation, a slightly different problem.
More recently, \cite{posecnn} decouple or `disentangle' translation and rotation terms by moving the centre of rotation to the centre of the object and representing translation in 2D image space.
In practice, regression to 3D orientation has had limited success \cite{regression,Sundermeyer}.
This is partly due to such methods only covering the small subsection of pose space that is seen during training \cite{ssd6d}, which has in turn led to discretisation of the pose space as discussed below.
Recently, \cite{self6d} obtained state-of-the-art results on real data by separately regressing rotation and translation within a self-supervised framework that also utlises RGBD images.

\textbf{Pose Classification:}
Pose classification has typically involved discretising $SO(3)$, while the translation component is obtained via regression.
However, as even a coarse discretisation of $SO(3)$ ($\sim5^\circ$ precision) leads to over 50,000 possible cases \cite{Sundermeyer}.
Kehl \textit{et al.} \cite{ssd6d} approach this problem by decomposing 6D pose into viewpoint and discretised in-plane rotation.
However, as noted in \cite{Sundermeyer}, this approach leads to ambiguous classifications as a change of viewpoint can be nearly equivalent to a change of in-plane rotation.
Furthermore, such classification leads to coarse pose estimates that require further refinement \cite{ssd6d,deepim}, as discussed below.

\textbf{Pose via an Intermediate Representation:}
A recent, popular approach to pose estimation is to first regress to an intermediate representation such as keypoints, from which pose can be obtained via 2D-3D correspondences and a PnP algorithm \cite{tekin,bb8,pvnet,hybridpose,dpod,cdpn,epos}.
Of these approaches, some, including \cite{tekin,bb8} regress to a set of bounding box corners.
Such methods encounter difficulty when objects are occluded or truncated.
Alternatively, \cite{pvnet} predicts pixel-wise unit-vectors that in turn indicate direction to keypoints.
Song \textit{et al.} \cite{hybridpose} implements these vector directions from \cite{pvnet} while also estimating object edge vectors and symmetry correspondences.
Zakharov \textit{et al.} \cite{dpod} apply texture to objects with a 2D image generated from spherical or cylindrical projections, and uses these to estimate dense correspondences.
Li \textit{et al.} \cite{cdpn} apply this intermediate representation approach to estimate rotation, but estimates translation separately via regression.

\subsection{Pose Refinement} 
Of the networks discussed above, several have additional refinement steps that can be added to improve results.
For example, DeepIM \cite{deepim} is presented as a refinement step for POSECNN \cite{posecnn}, the work \cite{manhardt} is presented as a refinement step for SSD6D \cite{ssd6d}, and DPOD \cite{dpod} is presented along with a refinement step.
In each case a synthetic image of the target object is rendered using the initial pose estimate and a 3D model of the target object.

The key difference in these approaches is in the representation of the relative pose.
DeepIM \cite{deepim} utilise an `untangled representation' in which the centre of the object is the centre of rotation, and translation is estimated in pixel space.
CosyPose \cite{cosypose} improve upon DeepIM by utilising a more recent feature detection network, and implement their framework to estimate object poses in a multi-view scene.
Manhardt \textit{et al.} \cite{manhardt} represent rotation as a unit quaternion and translation as a vector in $\mathbb{R}^3$.
DPOD \cite{dpod} combines these approaches by estimating rotation relative to the object centre, and estimating translation as a vector in $\mathbb{R}^3$.
These approaches therefore differ mainly in their loss functions, and all require textured 3D object models.

\section{Proposed Innovation CNN \label{approach}}
In this section we present a formulation for a general state estimation algorithm and apply this formulation to the problem of object pose estimation.
We also present our network architecture, training loss, and network implementation.

\subsection{State Estimation}
Motivated by filtering and optimisation principles \cite{anderson}, we treat object pose estimation as an offline state estimation task.
In this context a \textit{state} is an internal representation of a system that is sufficient to fully define the future evolution of all system variables.
A \textit{state estimation} algorithm is a dynamical system that takes measurements from the true system as inputs and whose state is an estimate of the true system state.

Consider a dynamical system in a state $\bm{X}$ defined by
\begin{align}
    \mathbf{\dot{X}}(t) &=f(\mathbf{X}(t),\mathbf{V}(t)), \\
    \mathbf{y}(t) &= h(\mathbf{X}(t)),
\end{align}
where $f(.)$ is the state model $\textbf{V}(t)$ is an measured input signal, $h: \bm{X}(t) \to \mathbf{y}(t)$ is an output map, and $t$ represents the current time.
We think of the output $\mathbf{y}(t)$ as encoding the specific target information that the engineer is interested in, while the state $\mathbf{X}$ encodes the full information necessary for the propagation of the dynamical system.

The class of estimation algorithm considered in this paper are of the form
\begin{align}
    \mathbf{\dot{\widehat{X}}}(t) &=f(\mathbf{\widehat{X}}(t),\mathbf{V}(t))  - \Delta(t), \\
    \mathbf{\widehat{y}}(t) &= h(\mathbf{\widehat{X}}(t)),
\end{align}
where $\Delta$ is the innovation term that is chosen so as to drive the estimated outputs towards the true outputs, and $\widehat{\cdot}$ represents an estimation.

\subsection{Object Pose Refinement via State Estimation\label{application}}
The proposed approach involves taking the output state from an existing object pose estimation network and refining this output using a state estimation framework to obtain more accurate pose estimates.
We choose PVNet \cite{pvnet} to be the baseline object pose estimation network.
In established algorithms such as PVNet, the state is estimated directly via a deep neural network.
A typical formulation for such a problem can be written
\begin{align}
\mathbf{\widehat{X}} &= \mathcal{E} (\mathbf{V}) \\
\mathbf{\widehat{y}} &= h(\mathbf{\widehat{X}}),
\end{align}
where the estimator $\mathcal{E}$ is implemented as a CNN or more classical computer vision algorithm.

The output of PVNet is a unit vector field, with vectors pointing from each pixel to each of a set of keypoints.
The vectors are defined to be
\begin{align}
    \bm{\eta}_{ij}^k &= \bm{\xi}^k-\bm{\xi}_{ij},
\end{align}
where $\bm{\xi}_{ij}$ denotes a 2D pixel with coordinates $(i,j)$ within an image $\mathcal{I}$ with dimensions $M\times N$,
and $\bm{\xi}^k$ represents the pixel location of keypoint $k\in\mathcal{K}$.
The unit vector field is
\begin{align}
    \bm{X}_{ij}^k & = \frac{\bm{\eta}_{ij}^k}{\|\bm{\eta}_{ij}^k\|_2}\in \mathbb{R}^{2\times \mathcal{K}\times M\times N}.
\end{align}
This unit vector field forms the state for the estimation problem.

We assume a static state model
\begin{align}
    f(\mathbf{X}(t),\mathbf{V}(t))  = 0. 
\end{align}
Approximating the time derivative by $\mathbf{\dot{X}} = \delta t(\mathbf{X}(t) - \mathbf{X}(t-1))$, the corresponding state estimation algorithm has the form
\begin{align}
    \mathbf{\widehat{X}}(t) &=\mathbf{\widehat{X}}(t-1) - \Delta(t-1) ,\\
    \mathbf{\widehat{y}}(t) &= h(\mathbf{\widehat{X}}(t)),
\end{align}
where $\delta t$ is absorbed into the innovation $\Delta$.

The output of the state estimation algorithm $\mathbf{\widehat{y}}$ is the object pose.
This is a discrete system with time steps $t\in\{0,1,...,T\}$.

The function $h$ maps a vector field representation of keypoints to object pose via a RANSAC and uncertainty-driven PnP framework (EPnP), as discussed in \cite{pvnet}.
The task then becomes to learn an appropriate $\Delta$, such that the error in the state estimation algorithm is iteratively driven towards zero.

Consider the standard L2-norm loss function
\begin{align}
    \Phi(t) &= \frac{1}{2}\|\widehat{\mathbf{X}}_{ij}^k(t) - \mathbf{\mathbf{X}}_{ij}^k\|^2_2. 
\end{align}
The gradient of this function is
\begin{align}
    \mathbf{\nabla} \Phi(t) &= \frac{1}{2}\nabla_{\mathbf{\widehat{X}}_{ij}^k(t)}\| \widehat{\mathbf{X}}_{ij}^k(t) - \mathbf{\mathbf{X}}_{ij}^k\|_2^2 \nonumber\\
    &= \mathbf{\widehat{X}}_{ij}^k(t) - \mathbf{X}_{ij}^k,\label{eq:gradient}
\end{align}
where $\mathbf{X}_{ij}^k$ is the ground truth vector field, and $\nabla_{x}$ denotes the gradient operator with respect to $x$.
Set
\begin{align}
    \Delta(t) = \alpha(t) \mathbf{\nabla} \Phi(t),
    \label{eq:grad}
\end{align}
where $\alpha(t)$ is a sequence of step-sizes that must be specified.

In practice, the gradient term $\nabla \Phi(t)$ is not directly measured. 
Instead, we use an estimate of this value,
\[
\widehat{\nabla \Phi}(t) = \mathcal{F}(\mathbf{V},\hat{\mathbf{X}}(t))
\]
where $\mathcal{F}$ represents the output of the CNN described in Section \ref{network}.

The iterative update is applied by employing gradient descent
\begin{align}
    \bm{\widehat{X}}_{ij}^k(t+1) &= \bm{\widehat{X}}_{ij}^k(t) - \alpha(t) \widehat{\nabla\Phi}(t). \label{eq:evaluation}
\end{align}
Within this iterative framework the initial state $\mathbf{\widehat{X}}_{ij}^k(0)$ is the vector field estimate obtained from PVNet, and $\widehat{\nabla\Phi}(t)$ is estimated by our network.
Algorithm \ref{alg:iteration} summarises the proposed information pipeline.

\begin{algorithm}[H]
    \caption{Iterative Optimisation with Innovation CNN}
    \label{alg:iteration}
    \begin{algorithmic}[1]
        \State Choose $0<\alpha(t)<1$
        \State Choose $T>0$ \Comment{Maximum \#iterations}
        \State Input: $I$, $\mathbf{P}^k$ \Comment{Image, 3D object points}
        \State $\bm{\widehat{X}}_{ij}^k(0) \leftarrow$ \text{ PVNet}($I$)
        \For{$t=1\to T$}
            \State $\widehat{\mathbf{\nabla}\Phi}(t) \leftarrow $\text{ Innov}($I,\bm{\widehat{X}}_{ij}^k(t)$)
            \State $\bm{\widehat{X}}_{ij}^k(t+1) = \bm{\widehat{X}}_{ij}^k(t) - \alpha(t) \widehat{\mathbf{\nabla}\Phi}(t)$
        \EndFor
        \State $\mathbf{\widehat{y}}$ = EPnP($\bm{\widehat{X}}_{ij}^k(T),\mathbf{P}^k$)
        \State Output: $\mathbf{\widehat{y}}$\Comment{Pose}
    \end{algorithmic}
\end{algorithm}

\begin{figure}[H]
    \centering
    \includegraphics[width=1\linewidth]{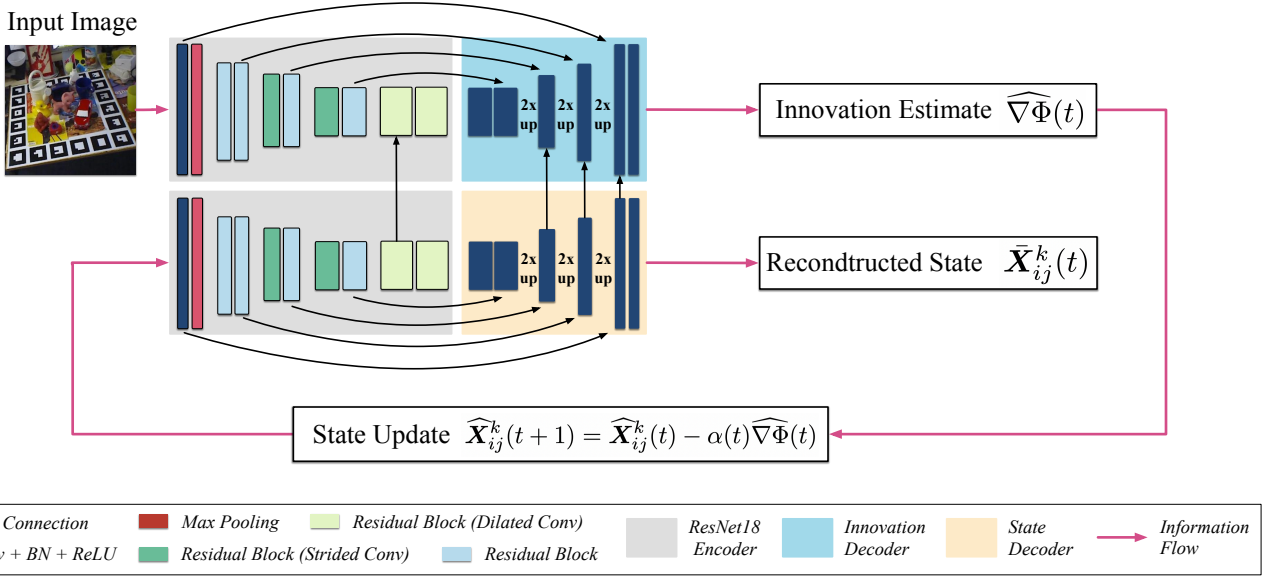}
    \caption{Network Architecture: Innovation Estimator (top block), and Estimate Autoencoder (bottom block).
    The input image (top left) is passed to the Innovation Estimator, which estimates $\widehat{\nabla\Phi}(t)$ (top right).
    The current state estimate $\widehat{\bm{X}}^k_{ij}(t)$ is passed to the State Encoder (bottom left).
    The output of the State Decoder (bottom right) is $\bar{\bm{X}}^k_{ij}(t)$, which represents the same information as $\widehat{\bm{X}}^k_{ij}(t)$.
    Using $\widehat{\nabla\Phi}(t)$ the state $\widehat{\bm{X}}^k_{ij}$ is updated via gradient descent (Eq. (\ref{eq:evaluation})).
    The updated state estimate $\widehat{\bm{X}}^k_{ij}(t+1)$ is then passed to the State Decoder in an iterative process.}
	\label{fig:arch}
\end{figure}

\subsection{Network Architecture \label{network}}
The network consists of two encoder-decoder blocks operating in parallel (see Figure \ref{fig:arch}).
The top block in Figure \ref{fig:arch} is labeled the \textit{Innovation Estimator}, and the bottom block is labeled the \textit{Estimate Autoencoder}.
The Innovation Encoder consists of a pre-trained ResNet-18 \cite{resnet} architecture.
The Innovation Decoder consists of successive convolution and upsampling operations.
The Estimate Autoencoder consists of a pre-trained ResNet-18 backbone followed by three successive upsampling layers.
The input dimensionality of the Estimate Autoencoder modified to accept a vector field state estimate.

The Innovation Estimator receives an image and returns an estimate of the gradient of the vector field state $\widehat{\mathbf{\nabla}\Phi}$.
The Estimate Autoencoder receives a vector field state $\bm{\widehat{X}}_{ij}^k$ and returns an estimate of the vector field state $\bar{\bm{X}}_{ij}^k$.
The Estimate Autoencoder is therefore an auxiliary task aimed at allowing information from the initial state estimate to be injected into the Innovation Estimator via skip connections.
This auxiliary task leads to improvement on the main task by leveraging domain specific information via multi-task learning.

\subsection{Training Loss}

The state loss is applied to the Esimate Autoencoder, and is defined by
\begin{align}
    \mathcal{L}_{\text{state}} &= \sum^\mathcal{K}_{k=1}\sum_{(i,j)\in \mathcal{S}}\|\widehat{\bm{X}}_{ij}^k(t) - \bar{\bm{X}}^k_{ij}(t)\|^2_1,
\end{align}
where $(i,j)\in\mathcal{S}$ indicates that the pixel is within the segmentation mask,
and the norm that is used is the `smooth l1 norm' for unit vectors proposed by \cite{fastrcnn}.
The state $\bar{\bm{X}}^k_{ij}$ represents the same information as $\widehat{\bm{X}}^k_{ij}$ and is used only for the purpose of training the autoencoder.

The state gradient loss is applied to the Innovation Estimator, and is defined by
\begin{align}
    \mathcal{L}_{\text{innov}} &= \sum^\mathcal{K}_{k=1}\sum_{(i,j)\in \mathcal{S}}\|\widehat{\nabla\Phi}(t)  - \nabla\Phi(t)\|^2_1\\
     &= \sum^\mathcal{K}_{k=1}\sum_{(i,j)\in \mathcal{S}}\|\widehat{\nabla\Phi}(t)  - (\widehat{\bm{X}}_{ij}^k(t) - \bm{X}^k_{ij})\|^2_1 .
\end{align}
The loss is backpropogated after each iteration of the gradient descent defined by Eq. (\ref{eq:evaluation}).

The loss function used to train the network is
\begin{align}
\mathcal{L} &= \mathcal{L}_{\text{innov}} + \gamma\mathcal{L}_{\text{state}}, \label{eq:loss}
\end{align}
where $\gamma$ is a scaling factor.

\subsection{Implementation Details}
The network is trained for $T=2$ iterations of Eq. (\ref{eq:evaluation}) at each epoch, with a constant step size of $\alpha=0.6$.
We use a pretrained PVNet \cite{pvnet} to provide the initial estimate $\bm{X}^k_{ij}(0)$ for each object.
We use a batch size of 64, and down-sample training images by four to fit the learning algorithm on our machine.
We compute 8 keypoints via farthest point sampling, from which object pose is obtained via uncertainty-driven PnP \cite{pvnet}.
We also render 10,000 images and synthesis 10,000 images for each object.
We employ data augmentation in the form of random cropping, resizing, rotation, colour jittering.
An Adam optimizer \cite{adam} is employed with an initial learning rate of 1$e^{-3}$, which is decayed to 1$e^{-5}$ by a factor of 0.85 every 10 epochs.
The learning hyperparameter in Eq. (\ref{eq:loss}) is set to $\gamma=10$.
We train our network for 50 epochs.

During testing, we employ Hough voting to localise keypoints, before using uncertainty-driven PnP to obtain object pose from keypoints.
We set the step size to $\alpha=0.01$ and perform iterations of Eq. (\ref{eq:gradient}) until the estimate converges.
Once the gradient output of our Innovation CNN approaches zero we consider the estimate to be converged.
Figure \ref{fig:add} provides qualitative results that illustrate this iterative convergence.

\section{Experiments \label{experiments}}
In this section we conduct experiments on two widely used datasets for object pose estimation, and evaluate our performance with two standard metrics for this task.

\subsection{Datasets}
We conduct experiments on two popular datasets for object pose estimation.

\textbf{LINEMOD} \cite{linemod} consists of 15783 images of 13 objects, with approximately 1200 instances for each object.
Challenging aspects of this dataset include lighting variations, scene clutter, object occlusions, and texture-less objects.

\textbf{Occlusion LINEMOD} \cite{occlusionlinemod} is a subset of LINEMOD images with each image containing multiple annotated objects under severe occlusion, thus presenting a more challenging scenario for accurate pose estimation.

\subsection{Evaluation Metrics}
We evaluate our approach using two standard metrics for object pose estimation.

The \textbf{ADD} metric \cite{linemod} computes the average 3D distance between the points of the 3D model under transformation from the ground truth and estimated pose respectively.
Specifically, given the ground truth rotation $\mathbf{R}$ and translation $\mathbf{T}$ and the estimated rotation $\mathbf{\hat{R}}$ and translation $\mathbf{\hat{T}}$, the ADD metric computes
\begin{align}
    \text{ADD} &= \frac{1}{m}\sum_{\mathbf{x}\in \mathcal{M}}\|\mathbf{(Rx+T) - (\hat{R}x+\hat{T})}\|_2, \label{eq:add}
\end{align}
where $\mathcal{M}$ denotes the set of 3D model points and $m$ is the number of points.
For symmetric objects we use the similar \textbf{ADD(-S)} metric \cite{posecnn}, where the mean distance is computed based on the closest point distance,
\begin{align}
    \text{ADD(S)} &= \frac{1}{m}\sum_{\mathbf{x}_1\in \mathcal{M}}\min_{\mathbf{x}_2\in \mathcal{M}}\|(\mathbf{Rx}_1+\mathbf{T}) - (\mathbf{\hat{R}x}_2+\mathbf{\hat{T}})\|_2. \label{eq:add}
\end{align}
We denote both metric as ADD(-S) and use the one appropriate to the object.
In each case a pose is considered correct if the average distance is less than 10\% of the 3D model diameter.
The reported metric is the percentage of poses that are considered correct.

The \textbf{2D projection} metric \cite{linemod} computes the mean distance between pixel locations of the 2D projections of the 3D model model, when projection is performed with the estimated and ground truth poses.
The metric is defined via
\begin{align}
    \text{2d Proj} &= \frac{1}{|V|}\sum_{\mathbf{v}\in V}\|\mathbf{PY}\mathbf{v}-\mathbf{P\widehat{Y}}\mathbf{v}\|_2, \label{eq:proj}
\end{align}
where $V$ is the set of all object model vertices, $\mathbf{Y}$ is the pose, and $\mathbf{P}$ is the camera matrix.
The estimated pose is considered correct if the average error is less than 5 pixels.
The metric reported is the percentage of correct poses.

Finally, we propose an additional metric, dubbed the \textit{State Distance} (SD), designed to quantify the difference between the estimated and ground truth vector field states.
This is used to provide an indication of whether the state estimate $\bm{\widehat{X}}^k_{ij}$ is converging to or diverging from the true state $\bm{X}^k_{ij}$.
This metric is defined by
\begin{equation}
    \text{SD} = \frac{1}{s}\sum_{(i,j)\in\mathcal{S}}\|\bm{\widehat{X}}^k_{ij}(t) - \bm{X}^k_{ij}\|^2_2, \label{eq:norm_x}
\end{equation}
where $(i,j)\in\mathcal{S}$ indicates that the pixel is within the segmentation mask, and $s$ denotes the number of pixels within the segmentation mask.

\subsection{Comparison to State-of-the-Art}
\begin{figure}[H]
    \centering
    \includegraphics[width=0.8\linewidth]{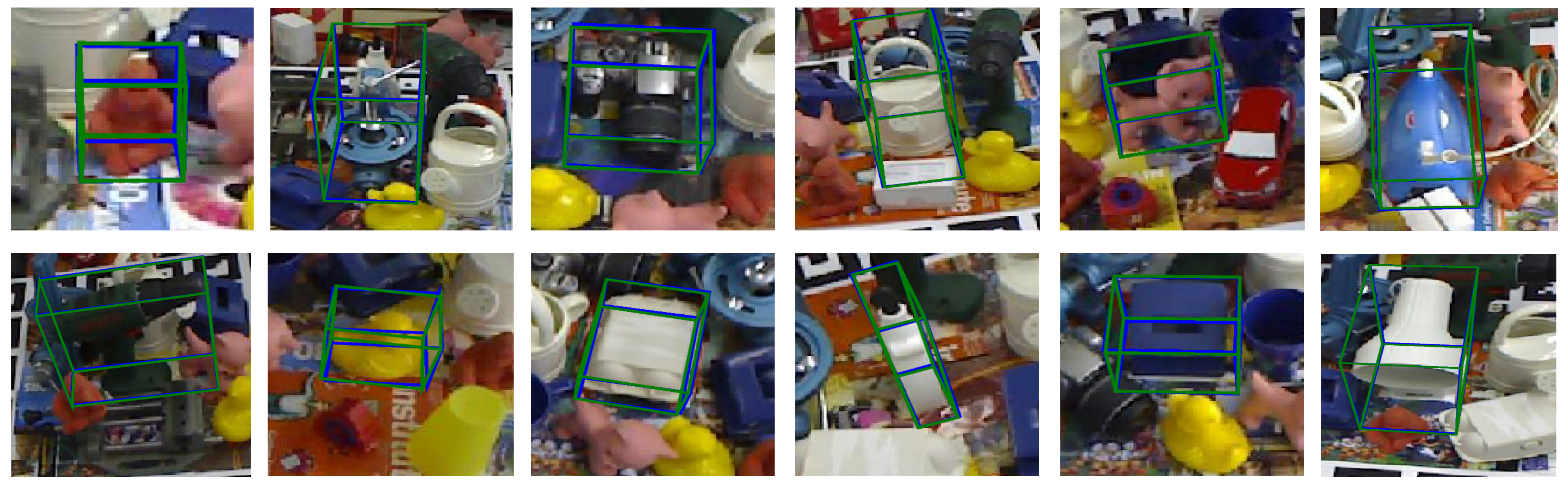}
    \caption{Visualisation of our qualitative results on the \textbf{LINEMOD} dataset.
    {\color{green}Green} bounding boxes represent ground truth poses and {\color{blue}blue} boxes represent our results.}
	\label{fig:linemod}
\end{figure}

\begin{figure}[H]
    \centering
    \includegraphics[width=0.5\linewidth]{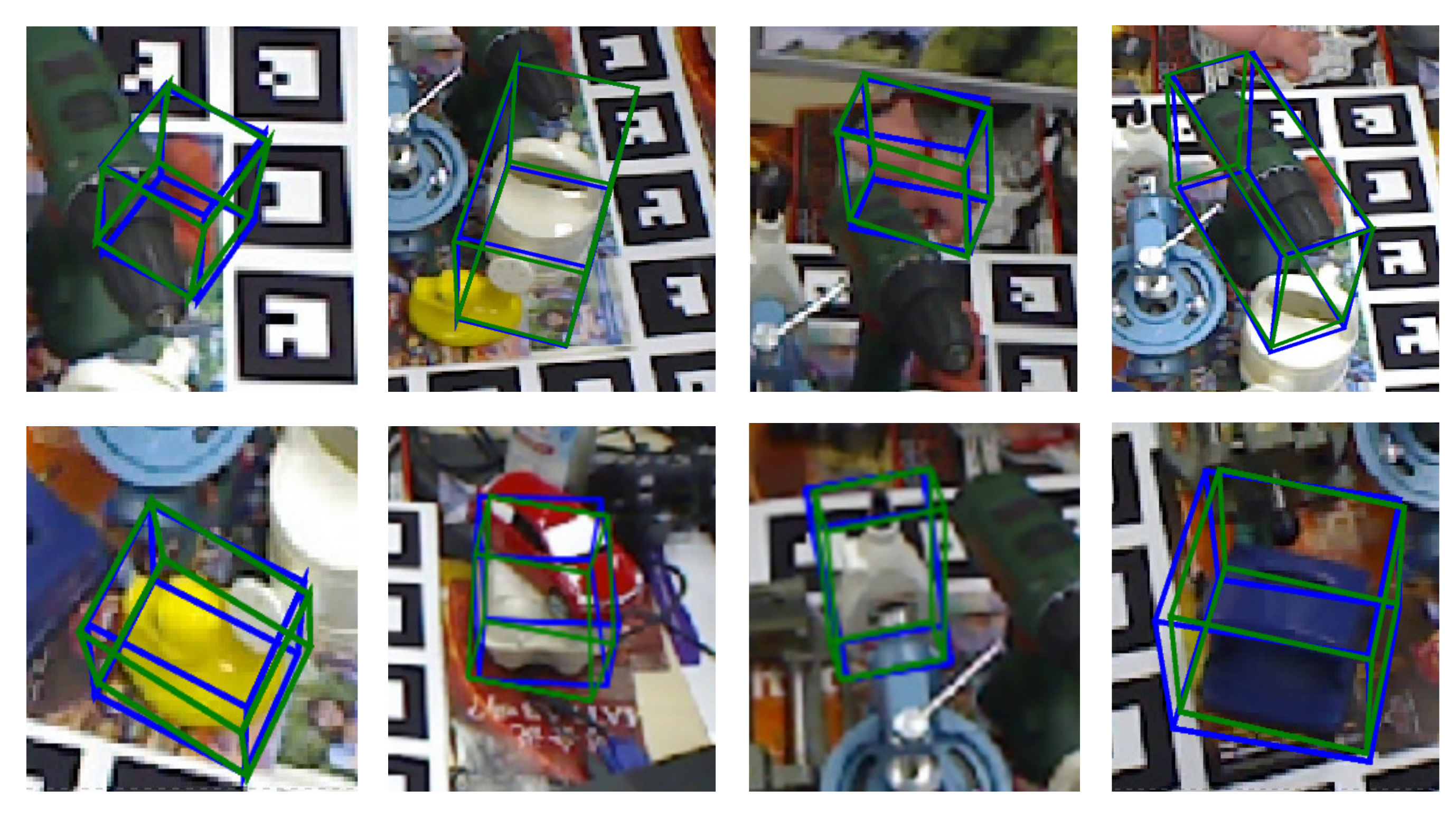}
    \caption{Visualisation of our qualitative results on each object of the \textbf{Occlusion LINEMOD} dataset.
    {\color{green}Green} bounding boxes represent ground truth poses and {\color{blue}blue} boxes represent our results.}
	\label{fig:occ_linemod}
\end{figure}

Results in terms of the ADD(-S) and 2D projection metrics on the LINEMOD and Occlusion LINEMOD datasets are presented in Tables \ref{tab:ADDcomp}, \ref{tab:2dproj}, and \ref{tab:ADDcompOcc}.
Qualitative results illustrating our final pose estimates are provided in Figures \ref{fig:linemod} and \ref{fig:occ_linemod}.

\begin{figure*}
    \centering
    \begin{subfigure}[b]{0.45\textwidth}
        \hspace{-0.4in}
        \includegraphics[width=1.3\textwidth]{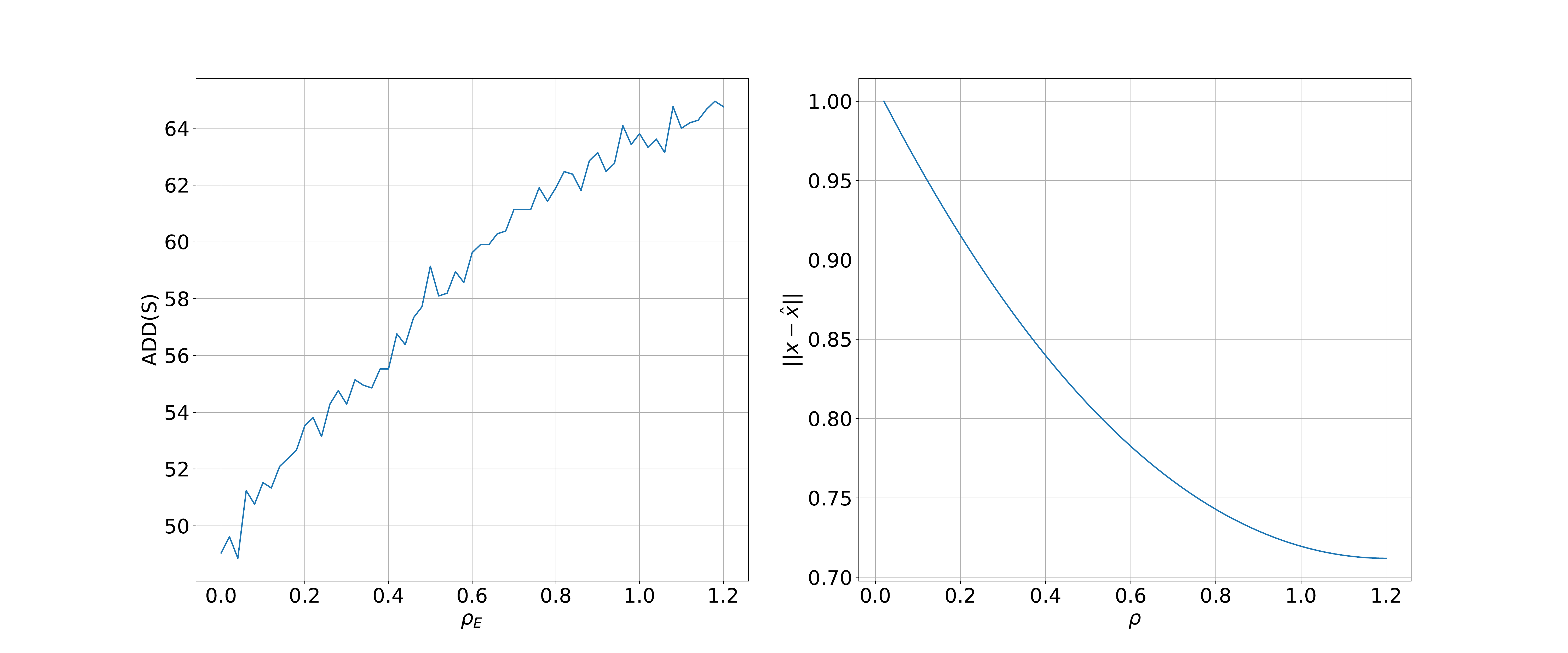}
        \caption{Ape, $\alpha=0.01$, $T=120$}
    \end{subfigure}
    \hfill
    \begin{subfigure}[b]{0.45\textwidth}
        \hspace{-0.5in}
        \includegraphics[width=1.3\textwidth]{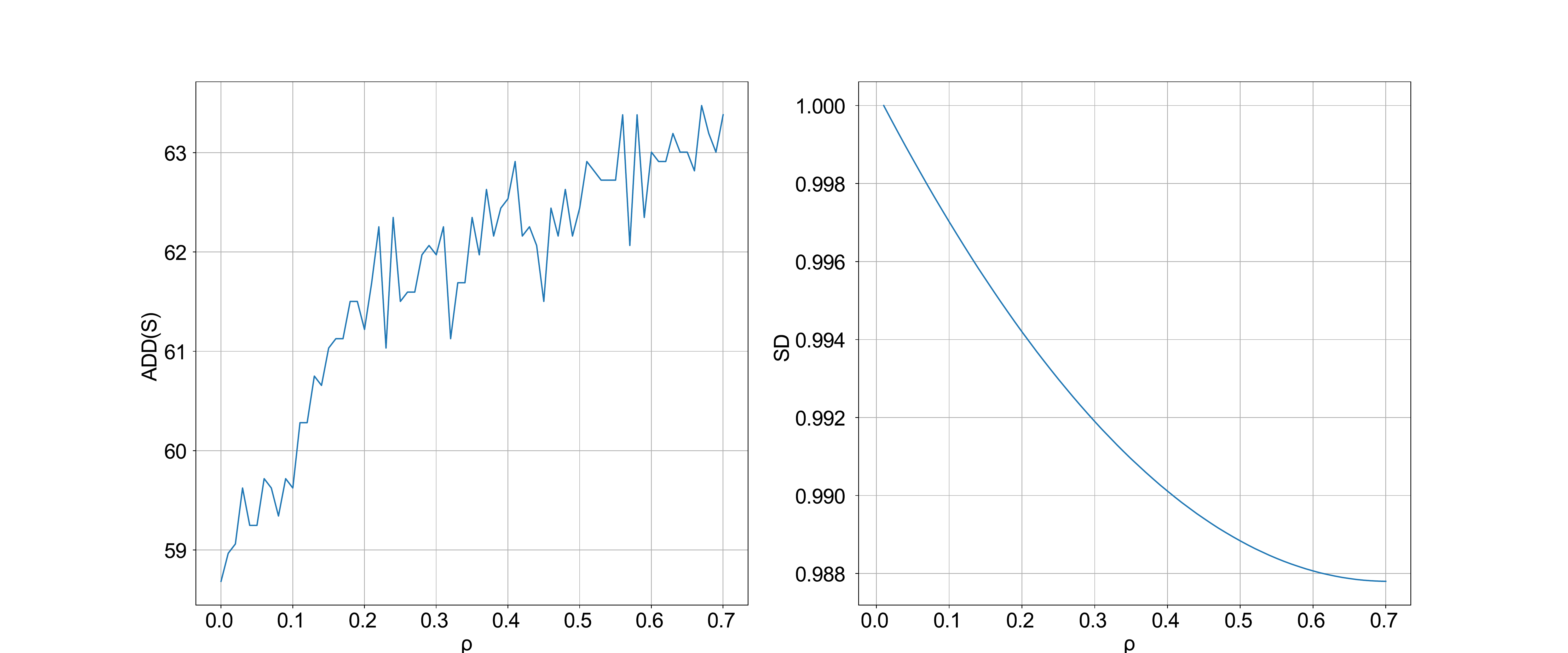}
        \caption{Duck, $\alpha=0.01$, $T=70$.}
    \end{subfigure}
    \caption{Iterative improvement on the \textbf{ADD(-S)} metric and corresponding percentage decrease in the State Distance (SD) for the ape and duck objects of the \textbf{LINEMOD} dataset
    plotted against the interpolation distance $\rho$ (Eq. \eqref{eq:rho}).}
    \label{fig:add}
\end{figure*}

\begin{table}[H]
    \centering
    \resizebox{0.6\columnwidth}{!}{%
    \begin{tabular}{c|cccc}
    \hline
    Methods            & \textbf{DPOD} \cite{dpod}& \textbf{CDPN} \cite{cdpn} &  \textbf{PVNet} \cite{pvnet} & \textbf{OURS} \\ \hline
    ape         & 53.28 & 64.38 & 43.62 & \textbf{64.76}           \\
    benchwise   & 95.34 & 97.77 & \textbf{99.90} &  \textbf{99.90}             \\
    cam         & 90.36 & 91.67 & 86.86 &  \textbf{92.58}          \\
    can         & 94.10 & 95.87 & 95.47 &  \textbf{96.56}           \\
    cat         & 60.38 & \textbf{83.83} & 79.34 &   82.83           \\
    driller     & \textbf{97.72} & 96.23 & 96.43 &   97.52            \\
    duck        & 66.01 & \textbf{66.76} & 52.58 &   63.31           \\
    eggbox      & \textbf{99.72} & 99.72 & 99.15 & 99.32            \\
    glue        & 93.83 & \textbf{99.61} & 95.66 &  96.91             \\
    holepuncher & 65.83 & \textbf{85.82} & 81.92 &  82.20           \\
    iron        & \textbf{99.80} & 97.85 & 98.88 &  99.31             \\
    lamp        & 88.11 & 97.86 & 99.33 &  \textbf{99.42}            \\
    phone       &  74.24 & 90.75 & 92.41 &   \textbf{94.33}        \\ \hline
    \textbf{average}& 82.98  & 89.86 & 86.27&   \textbf{89.92}           \\ \hline
    \end{tabular}
    }
    \caption{Performance comparison on the \textbf{LINEMOD} dataset with respect to the \textbf{ADD(-S)} metric.}
    \label{tab:ADDcomp}
\end{table}

\begin{table}[t]
    \centering
    \resizebox{0.6\columnwidth}{!}{%
    \begin{tabular}{c|cccc}
    \hline
     Methods           & \textbf{CDPN} \cite{cdpn}  & \textbf{YOLO6D} \cite{yolo}&\textbf{PVNet} \cite{pvnet} & \textbf{OURS} \\ \hline
    ape         &       92.10         &    96.86             &  \textbf{99.23}  &   \textbf{99.23}         \\
    benchwise   &      95.06         &    98.35             & \textbf{99.81}    &  \textbf{99.81}           \\
    cam         &      93.24         &    98.73             & \textbf{99.21}    &  99.12          \\
    can         &       97.44         &    99.41             & \textbf{99.90}   & 99.70            \\
    cat         &         97.41      &    99.80             & 99.30    &  \textbf{99.70}           \\
    driller     &          79.41      &    95.34             &  96.92  &  \textbf{97.23}           \\
    duck        &          94.65      &    \textbf{98.59}             & 98.02   &   98.02          \\
    eggbox      &       90.33      &    98.97             &  \textbf{99.34}     &  99.24           \\
    glue        &         96.53      &    \textbf{99.23}             & 98.45    &   98.45          \\
    holepuncher &          92.86       &    99.71             & \textbf{100.0}  &    99.70        \\
    iron        &         82.94      &    97.24             &  \textbf{99.18}   &  \textbf{99.18}           \\
    lamp        &    79.87          &    95.49             &  \textbf{98.27}   &   98.18        \\
    phone       &        86.07      &    97.64             &  \textbf{99.42}    &   \textbf{99.42}        \\ \hline
    \textbf{average}        &       90.37        &    98.10      &  \textbf{99.00}      &   \textbf{99.00}             \\ \hline
    \end{tabular}
    }
    \caption{Performance comparison on the \textbf{LINEMOD} dataset with respect to the \textbf{2D Projection} error.}
    \label{tab:2dproj}
\end{table}

\begin{table}[t]
    \centering
    \resizebox{0.6\columnwidth}{!}{%
    \begin{tabular}{c|cccc}
    \hline
     Methods           & \textbf{DPOD} \cite{dpod}  & \textbf{Pix2Pose} \cite{pix2pose} & \textbf{PVNet} \cite{pvnet}& \textbf{OURS} \\ \hline
    ape         & -  &22.0&15.81&\textbf{26.41}\\
    can         & -  &44.7&\textbf{63.30}&61.31\\
    cat         & -  &\textbf{22.7}&16.68&19.88\\
    driller     & -  &44.7&65.65&\textbf{70.10}\\
    duck        & -  &15.0&25.24&\textbf{31.99}\\
    eggbox      & -  &25.2&\textbf{50.17}&49.44\\
    glue        & -  &32.4&49.62&\textbf{51.16}\\
    holepuncher & -  &\textbf{49.5}&39.67&42.34\\ \hline
    \textbf{average}&32.79&32.0&40.77&\textbf{44.08}\\ \hline
    \end{tabular}
    }
    \caption{Performance comparison on the \textbf{Occlusion LINEMOD} dataset with respect to the \textbf{ADD(-S)} metric.}
    \label{tab:ADDcompOcc}
\end{table}

\begin{figure}[H]
    \centering
    \begin{subfigure}[b]{0.6\textwidth}
        \caption*{$0$\hspace{1.7cm}$T/3$\hspace{1.7cm}$T/6$\hspace{1.7cm}$T$}
        \hspace{-0.1in}
        \includegraphics[width=\textwidth]{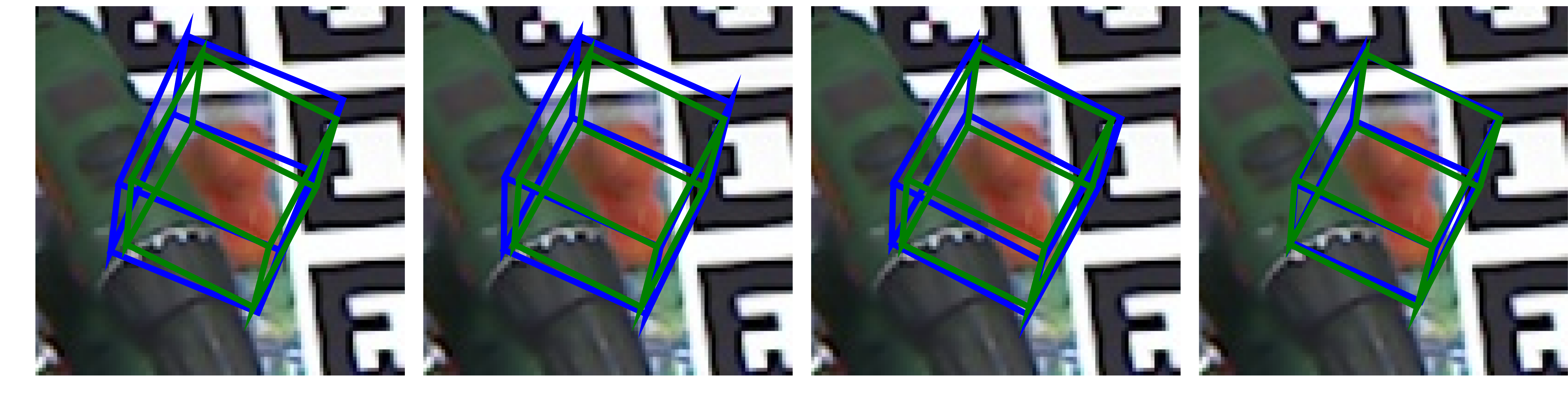}
    \end{subfigure}
    \hfill
    \begin{subfigure}[b]{0.6\textwidth}
        \hspace{-0.1in}
        \includegraphics[width=\textwidth]{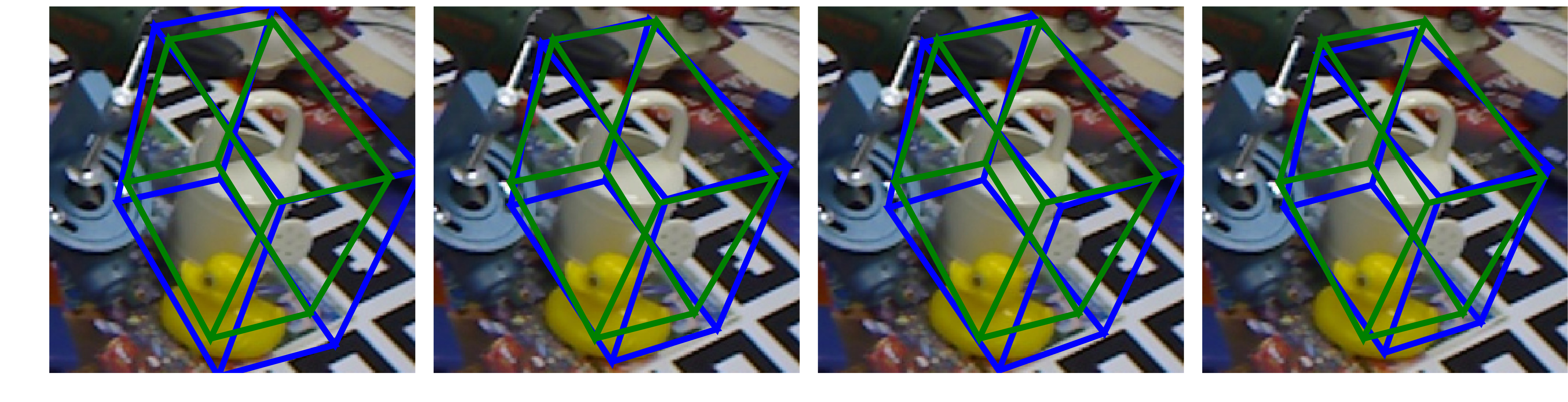}
    \end{subfigure}
    \hfill
    \begin{subfigure}[b]{0.6\textwidth}
        \hspace{-0.1in}
        \includegraphics[width=\textwidth]{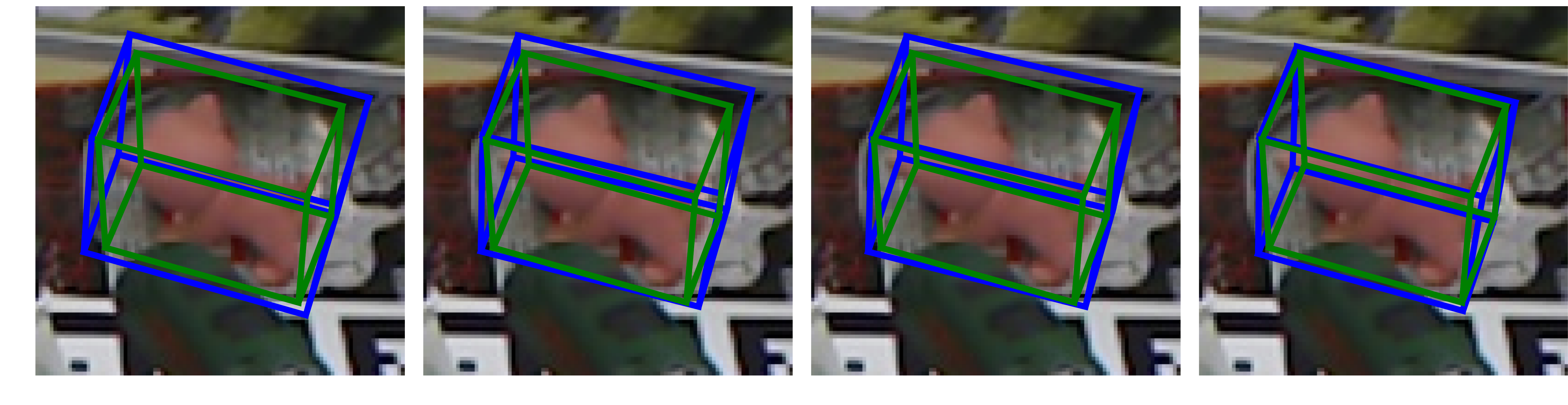}
    \end{subfigure}
    \hfill
    \begin{subfigure}[b]{0.6\textwidth}
        \hspace{-0.1in}
        \includegraphics[width=\textwidth]{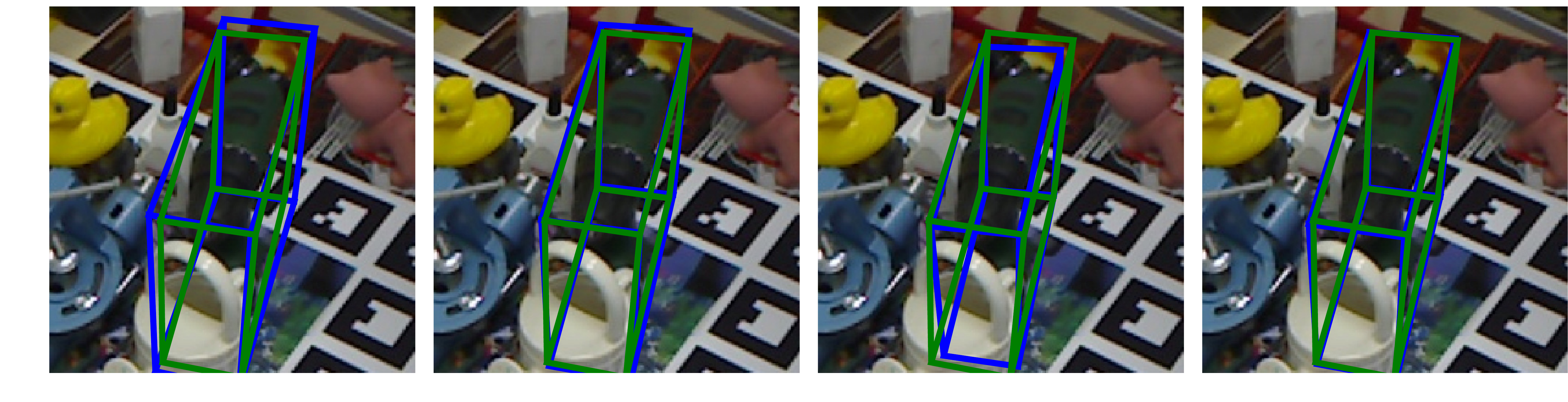}
    \end{subfigure}
    \hfill
    \begin{subfigure}[b]{0.6\textwidth}
        \hspace{-0.1in}
        \includegraphics[width=\textwidth]{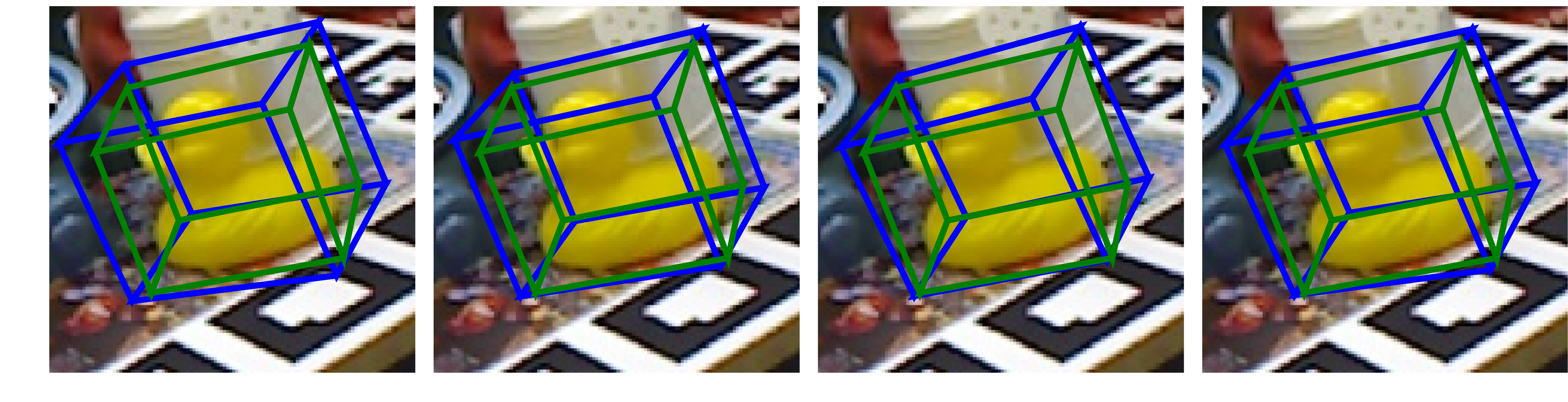}
    \end{subfigure}
    \hfill
    \begin{subfigure}[b]{0.6\textwidth}
        \hspace{-0.1in}
        \includegraphics[width=\textwidth]{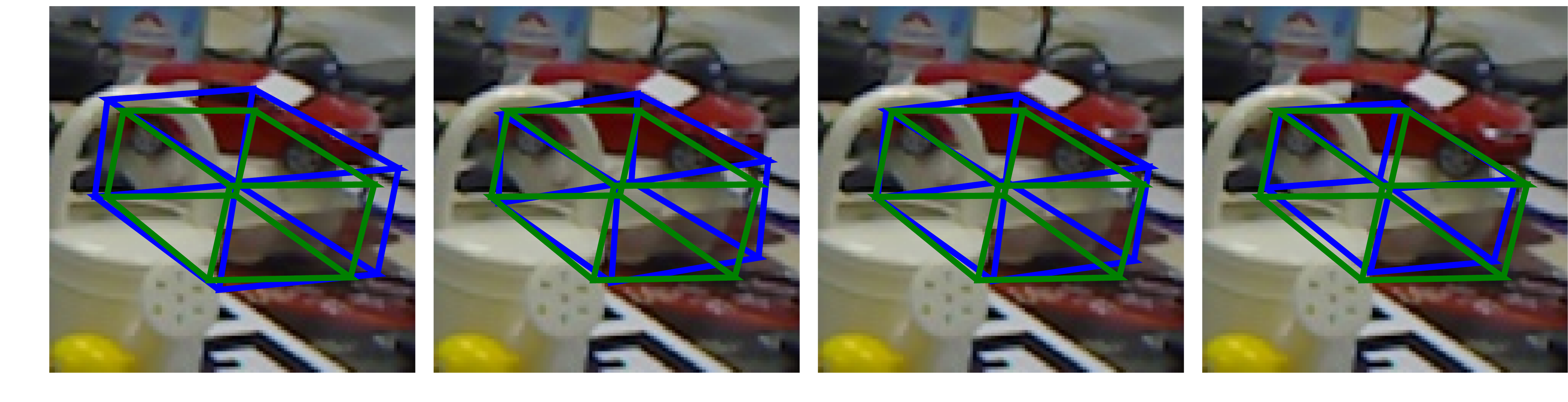}
    \end{subfigure}
    \hfill
    \begin{subfigure}[b]{0.6\textwidth}
        \hspace{-0.1in}
        \includegraphics[width=\textwidth]{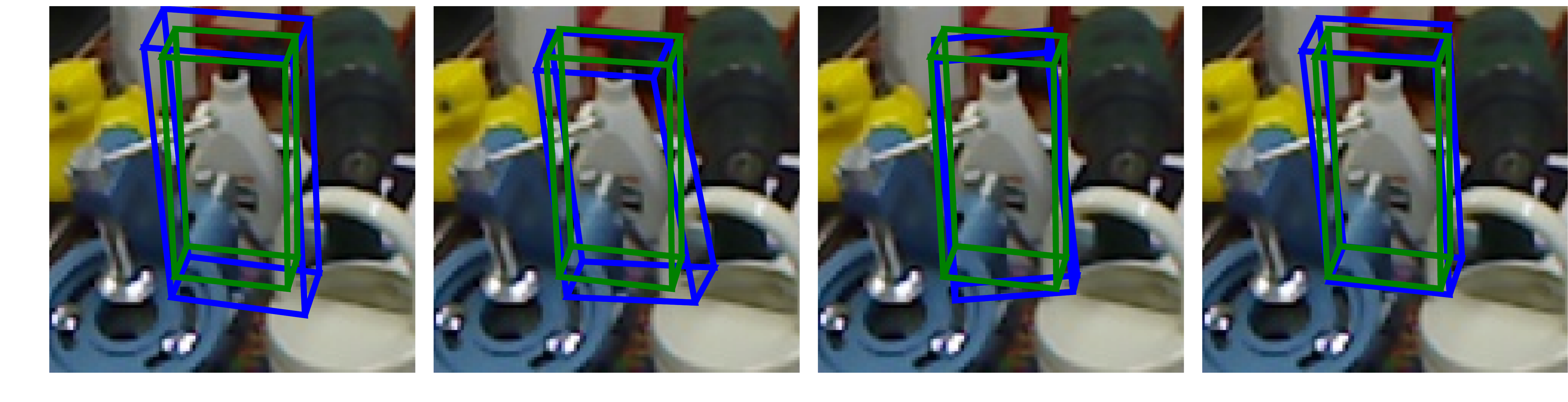}
    \end{subfigure}
    \hfill
    \begin{subfigure}[b]{0.6\textwidth}
        \hspace{-0.1in}
        \includegraphics[width=\textwidth]{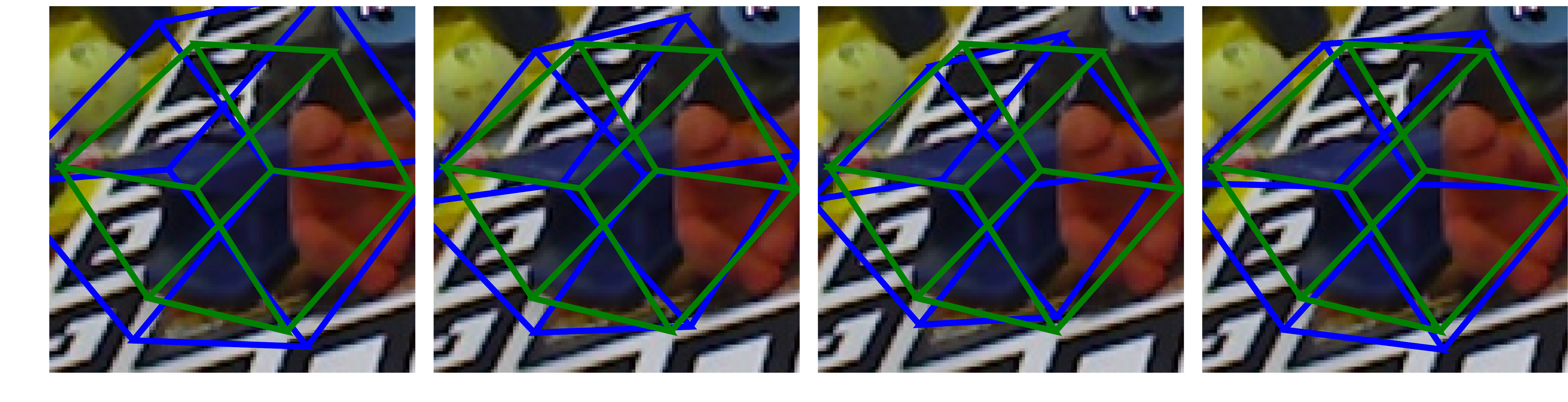}
    \end{subfigure}
    \caption{Iterative pose refinement on all objects in the \textbf{Occlusion LINEMOD} dataset.
    The initial pose estimate is shown in the left ($t=0$), and the final pose estimate is shown on the right ($t=T$), with equidistant intermediate poses shown in-between.
    {\color{green}Green} bounding boxes represent ground truth poses and {\color{blue}blue} boxes represent our results.
    }
       \label{fig:iter}
\end{figure}

In this paper we focus on refining an initial pose estimate without leveraging any extra supervision.
In Tables \ref{tab:ADDcomp},\ref{tab:2dproj},\ref{tab:ADDcompOcc} we compare our results to the top three other pose estimation networks in this category.
Our approach outperforms all previous methods and achieves state-of-the-art performance on both the LINEMOD and Occlusion LINEMOD datasets.

Most significantly, our network generates pose estimates that lead to a higher average on both the ADD(-S) and 2D projection metrics, compared to the baseline network.
This includes a performance increase of 3.69\% on the LINEMOD dataset and 3.31\% on the Occlusion LINEMOD dataset in terms of the ADD(-S) metric.
This is equivalent to improving the performance by 4.2\% and 8.12\% respectively compared to the performance of the baseline network.

It can be seen that the largest improvement is obtained for objects for which the baseline network provides a relatively poor initial estimate.
Such objects include the relatively texture-less `ape' and `duck' and all objects in the Occlusion LINEMOD dataset.
Specifically, on the ape and duck objects we improve baseline estimates by 21.14\% and 10.73\% respectively in terms of the ADD(-S) metric.
Qualitative results for these two objects are presented in Figure \ref{fig:add}, for which the ADD(-S) metric is evaluated at each iteration of Eq. (\ref{eq:evaluation}).
Figure \ref{fig:add} also illustrates the iterative convergence of the state distance SD.
In this figure, the ADD(S) and SD metrics are plotted as a function of the \textit{interpolation distance},
\begin{align}
    \rho = \alpha(t) T.
    \label{eq:rho}
\end{align}
The interpolation distance quantifies how far we have iterated from the initial estimate.

Qualitative results illustrating the iterative application of our method are presented for the Occlusion LINEMOD dataset in Figure \ref{fig:iter}.
To generate Figure \ref{fig:iter} the initial and final pose are displayed, along with two intermediate poses sampled from the iterative framework.
For example, for the Ape object $T=120$ iterations were used, and Figure \ref{fig:iter} displays poses for T=\{0,40,80,120\}.

\subsection{Ablation Study}

We conducted experiments with a range of configurations before choosing the configuration used to produce the presented results.
Results of these experiments are summarised in Table \ref{tab:ablation}.
In Table \ref{tab:ablation}, column one shows a comparison of training with (\cmark) and without (\xmark) the Estimate Autoencoder.
Column two shows a comparison of backpropograting each iteration of Eq. (\ref{eq:evaluation}) (\cmark) and backpropograting after $T$ iterations (\xmark).
Column three shows a comparison of choosing the output of PVNet as $\bm{X}^k_{ij}(0)$ (\cmark) and choosing the ground truth vector field computed from keypoints perturbed with 1\% noise (\xmark).
All ablation experiments were conducted on the Ape object of the \textbf{LINEMOD} dataset.

Results shown are the mean \textbf{ADD(-S)} results obtained from testing the final 20 epochs of the model.
In Table \ref{tab:ablation} we compare the results of experiments with and without the Estimate Autoencoder (the bottom network block in Figure \ref{fig:arch}).
We also compare our backpropgation scheme (backprop each iteration of Eq. (\ref{eq:evaluation})) with the alternative scheme of accumulating the loss and backpropogating after $T$ iterations, as discussed in \cite{irr}.
Finally we compare the impact of using a pretrained PVNet as the initial estimate $\bm{X}^k_{ij}(0)$ against using the ground truth vector field, or the ground truth vector field perturbed with different levels of noise.
We trialed a range of different noise levels, and the result of applying 1\% noise is presented in Table \ref{tab:ablation}.

\begin{table}[H]
    \centering
    \resizebox{0.5\columnwidth}{!}{%
    \begin{tabular}{l|ccc}
    \hline
     & \begin{tabular}[c]{@{}c@{}}Estimate\\ Autoencoder\end{tabular} & \begin{tabular}[c]{@{}c@{}}Backprop\\ each iteration\end{tabular} &  \begin{tabular}[c]{@{}c@{}} PVNet \\ Initial Estimate\end{tabular}\\ \hline
     \cmark &   \textbf{53.93}    &         52.01   & \textbf{53.93} \\
     \xmark &     50.61           &               \textbf{53.93}  & 46.00   \\ \hline
    \end{tabular}
    }
    \caption{Ablation Study of key design choices.
    The study indicates that the State Autoencoder improves performance, that it is better to backpropogate after each iteration of Eq. \eqref{eq:evaluation}, and that it is better to use the better to use the output of PVNet as the initial estimate during evaluation.}
    \label{tab:ablation}
\end{table}

\section{Conclusion \label{conclusion}}
In this paper we provide a new framework to improve an initial pose estimate.
We present a novel CNN architecture capable of estimating the gradient from an initial state prediction to the true state.
Our Innovation CNN refines initial state estimates in an SGD framework iteratively, thus greatly reducing the difficulty of estimating object poses in a single forward pass.
Extensive experiments on widely used datasets demonstrate that we improve initial object pose estimates significantly and obtain state-of-the-art performance.
Moreover, our Innovation CNN is generic but not restricted to our baseline method and we will apply it to different baseline networks in our future work.

\section*{Acknowledgements}
This research was supported in part by the Australian Government Research Training Program Scholarship and in part by the Australian Research Council through the ``Australian Centre of Excellence for Robotic Vision'' CE140100016.

{\small
\bibliographystyle{alpha}
\bibliography{egbib}
}



\end{document}